\documentclass{article}

\usepackage[hidelinks]{hyperref}

\usepackage[ruled]{algorithm2e}
\usepackage{graphicx}
\usepackage{float}
\floatstyle{boxed}
\restylefloat{table}

\usepackage[usenames,dvipsnames,table]{xcolor}
\usepackage{bchart}

\SetAlFnt{\small}
\SetAlCapFnt{\small}
\SetAlCapNameFnt{\small}
\SetAlCapHSkip{0pt}
\IncMargin{-\parindent}

\usepackage{soul}

\newcommand{\fif}[1]{{#1}} 
\newcommand{\doms}[1]{{#1}} 
\newcommand{\roy}[1]{{#1}} 

\newcommand\kompai{Kompa\"{i}}

\newcommand\cinter{\textit{\textsc{interact}}}
\newcommand\cwinter{\textit{\textsc{will\_interact}}}
\newcommand\clinter{\textit{\textsc{leave\_interact}}}

\newcommand\cnoone{\textit{\textsc{no\_one}}}
\newcommand\csomeone{\textit{\textsc{someone\_around}}}

\DeclareTextSymbol{\degre}{T1}{6}
\DeclareTextSymbol{\degre}{OT1}{23}

{}

\usepackage{tikz}
\usetikzlibrary{calc}
\usepackage[margin=2.8cm]{geometry}

\makeatletter
\def\hlinewd#1{%
\noalign{\ifnum0=`}\fi\hrule \@height #1 %
\futurelet\reserved@a\@xhline}
\makeatother

\begin{document}


\title{Starting Engagement Detection Toward a Companion Robot Using Multimodal Features\footnote{Author version.}} 

\author{{Dominique Vaufreydaz\textsuperscript{1,2}, Wafa Johal\textsuperscript{2}, Claudine Combe\textsuperscript{1}}\\
\small{\textsuperscript{1}Prima/Inria-LIG, CNRS}\\
\small{\textsuperscript{2}University of Grenoble - Alpes, LIG, CNRS}\\
}


\maketitle

\begin{abstract}
Recognition of intentions is a subconscious cognitive process vital to human communication.
This skill enables anticipation and increases the quality of interactions between humans.
Within the context of engagement, non-verbal signals are used to communicate the intention of starting the interaction with a partner.
In this paper, we investigated methods to detect these signals in order to allow a robot to know when it is about to be addressed.
Originality of our approach resides in taking inspiration from social and cognitive sciences to perform our perception task.
We investigate meaningful features, i.e. human readable features, and elicit which of these are important for recognizing someone's intention of starting an interaction.
Classically, spatial information like the human position and speed, the human-robot distance are used to detect the engagement.
Our approach integrates multimodal features gathered using a companion robot equipped with a Kinect.
The evaluation on our corpus collected in spontaneous conditions highlights its robustness and validates the use of such a technique in a real environment.
Experimental validation shows that multimodal features set gives better precision and recall than using only spatial and speed features.
We also demonstrate that 7 selected features are sufficient to provide a good starting engagement detection score.
In our last investigation, we show that among our full 99 features set, the space reduction is not a solved task.
This result opens new researches perspectives on multimodal engagement detection.
\end{abstract}

Keywords: multimodal perception - affective computing - healthcare technologies - companion robots



\section{Introduction}
\label{chap:introduction}

\begin{figure}[H]
 \begin{center}
 \includegraphics[width=0.6\linewidth]{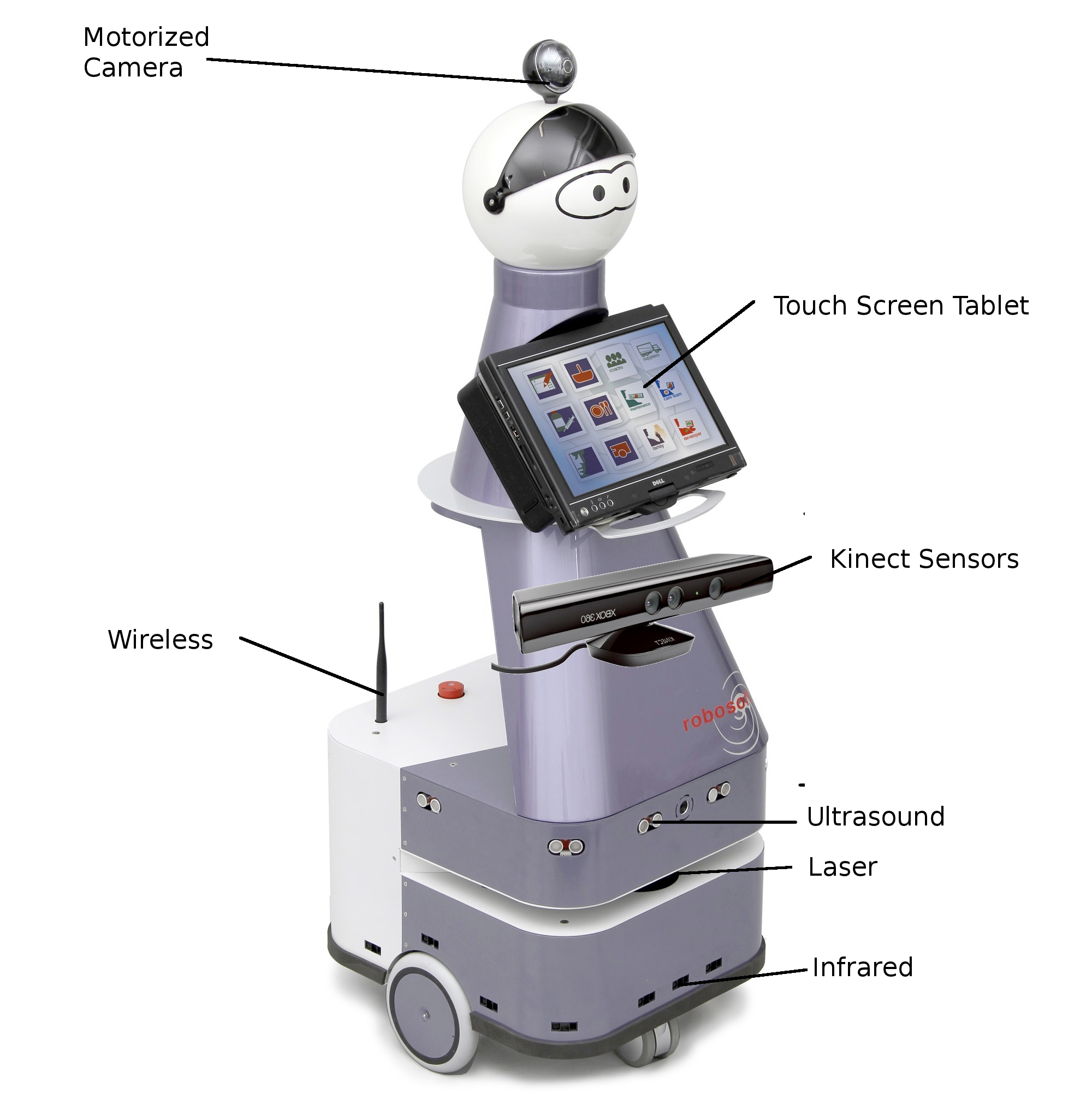}
\end{center}
\caption{The \kompai{} Robot from our partner Robosoft is equipped with a laser range finder, ultrasound and infrared telemeter, a tablet PC and a webcam on top. We added a Kinect for our experiments.}
\label{fig:kompaieq}
\end{figure}

\doms{
Companion robots are entities that are intended to be used as assistants in everyday life, those being personal coach, desktop manager, etc.
They could help to come up with tools that can potentially improve quality of life in the long run.
Among usual embedded functions, one can find entertainment, video conference, objects grasping, activity monitoring, serious games and frailty evaluation} \cite{ICRA20136630717, Fasola2013, fischinger2013hobbit, feil2005defining}.
Companion robots can assist therapy for autism \cite{castellano2009detecting}.
This paper presents research on companion robots using the Kompai Robot (see Figure \ref{fig:kompaieq})

As argued in \cite{kramer2011, pesty2011}, the primary challenge in building engaging
companion robots is to provide social competency in perceiving, reasoning and expressing
social and affective aspects of \roy{interactions} with the human user.
Companion robots \roy{are} aimed to interact with humans in home environments. 
In order to stay credible, companion robots are expected to behave and react \roy{as per predefined manners corresponding} to  instructions and social signals used by the user.

As speech being multimodal in a face-to-face interaction, non-verbal communication also uses a variety of channels to convey messages.
New areas explore techniques for the multimodal aspect of human communication in order to design robots able to read and express communicative signals in a social manner.
Non-verbal cues of communications have been well studied for detection of emotions \cite{Picard05, Zeng2009, Vinciarelli1920, Pantic2003}. In this paper, we propose to use these cues for intention recognition, and in particular \roy{an} intention of interaction. 

Recognition of intention is a basic skill acquired by infants early in their development.
According to Vernon \cite{Vernon2011}, among other skills, the perception of others' attention is crucial for the infant to master social interactions.
The perception of intentions and emotions, present in newborn infants, helps to set their \textquotedblleft preparedness\textquotedblright \ for social interaction \cite{Vernon2011}.
Intention recognition allows the interacting agent to take quick decisions and to respond better to the user's need or state of mind. 
Some of the non-verbal communication signals are cues to \roy{subdued} goals and intentions of the humans, and \roy{therefore} a good way to improve adaptability of the robots' behaviors is by predicting \roy{their} intentions.
A part of human cognition is anticipation, allowing \roy{reading} intentions and guessing goals in order to react quickly to stimuli.
This skill is also very important for turn-taking in interaction.

In neurocognition, the Broca's area, responsible for language comprehension, action recognition \& prediction and speech-associated gestures, would be the host of intention recognition in the human brain. 
According to Vernon, studies have shown that the activation of the Broca's area is significantly higher when a subject observes goal-directed actions with intentional cues rather than meaningless gestures.

As humans instinctively detect the intention of someone who wants to \roy{ask for way} in the street, we are interested in the \roy{opening} engagement phase \roy{of the process} during which humans \roy{subconciously} express their intentions to interact.
Our goal is to investigate techniques to detect and recognize signals for non-verbal communication reflecting this intention and in our particular case, the intention of a user to engage an interaction with a robot. 

Intention of engagement is a real question, especially when it comes to environments such as the work place or home, where people are not used to interact with robots \cite{Wang2010}.
Classically, the criterion for a user's intention of engagement is the spatial distance between the user and the communicant interface \cite{Glasnapp2009}. 
Some investigations have improved on this idea by also considering the speed of movement of the user \cite {Koo2009}. 
These studies have chosen to use \roy{the} relative spatial position \roy{of the concerned agents} as criteria. \roy{The following assumption is made behind this choice:} if the user is close to the robot, \roy{there is an intention to interact}.
Using distance and sometimes speed of the human \roy{provides with satisfactory} results, but for a companion robot in real \roy{situations} at home, close distance does not \roy{necessarily} signal a desire for engagement. 
For instance, many times during the day, one can pass in front of the refrigerator without the wish to open it.
Following the same logic, \roy{despite the physical distance of the user from the robot,} a robot \roy{should be able to} detect when it is about to be solicited, and anticipate the interaction in order to be more comfortable and socially acceptable.

\doms{In this study, we propose a multimodal approach for detecting a starting engagement using a RGB-D sensor mounted on a companion robot.
Getting inspiration from social and cognitive sciences, our goal is to select features in order to improve the re-usability in other situations and/or with other sensors.
In our approach, the idea here is to get rid of the usual way to do such experiment i.e. putting all available features together, combining them in a more optimized representation and let the training paradigm filter everything.
Doing this, we might have good performances, but we may not learn anything about detecting intention of engagement.
We will see that less than 10\% of our features are crucial for starting engagement detection.
In another context, one can make well-founded choices among sensors to reflect this knowledge.
It will be more efficient to design a new device or robot knowing which particular features are of importance.
This prospective research aim to build a set of meaningful features extracted from multimodal sensors useful for the description, recognition and discrimination of the intention of engagement.

This paper aims to contribute on the following statements : 
\begin{itemize}
\item There exist subconscious social signals expressed by humans that characterize their will to interact with a robot and these signals are detectable. 
\item Some features from literature in the social and cognitive sciences are computable on a companion robot (notably Schegloff metrics \cite{Schegloff1998}).
\item Multi-modal features will perform better than spatial features to detect this starting of engagement in a home-like environment. 
A realistic dataset in a home-like environment can help us to validate this hypothesis.
\item The set of relevant features for starting of interaction detection can be reduce without loss of performance using a feature space reduction process using the Minimum Redundancy Maximum Relevance (MRMR) method \cite{MRMR05} never used in this context.
\end{itemize}

}

\section{Multimodal Social Signal Processing For Non-Verbal Communication}
\label{chap:related-work}

\subsection{Social Signal Processing}
\label{sec:ssp}

A \roy{communicative} agent does not use only the verbal channel, but many channels to send and receive various messages while interacting \cite{Kaminski02}: human communication is intrinsically multimodal. 
To make human-robot communication fluent and acceptable, the robot has to decode these behavioral and non-verbal cues in order to act accordingly.
For instance, computer systems and devices able to recognize agreement or inattention, and capable of adapting and responding in real-time to these social signals in a polite, non-intrusive or persuasive manner, are likely to be perceived as more natural, efficient, and trustworthy\cite{SSPNet, Vernon2011}.
In the context of people assistance, social features seem to be crucial for \roy{the} acceptance of a robotic companion in a domestic environment.

Argyle\roy{,} in his book ``Bodily Communication'' \cite{argyle75} \roy{mentions} different signals from different modalities used for non-verbal communication. 
The considered modalities are facial \roy{expressions}, gaze, gestures \& body movements, posture, contact, spatial behavior, clothing, and vocalizations. 
This work shows that recording of these modalities allows to recognize the mood of a person.

P. E. Bull \cite{bull87} follows the idea that communication implies a socially shared signal system or code.
Non verbal communication is argued to be intentional or non-intentional.
Therefore, it is valuable information that allows to access intentions of the emitter that can be non-voluntarily transmitted. 
Bull claims the importance of posture and gesture in non verbal communication where these channels have been neglected compared to facial features and speech cues.

\subsection{Intentionality in Human-Machine Interaction}
\label{ssec:engage}

The intention cues form a way of communication. 
As stated before, recognition of human{'}s intentions, goals and actions is important in the improvement of non-verbal human-robot cooperation. 
\cite{Krauthausen2010} defines intention recognition as the process of estimating the force driving humans\roy{'} actions based on noisy observations of human\roy{'} interaction with \roy{their} environment. 
Tahboub in \cite{Tahboub2006} sees intention recognition as a substitution or a complement to reliable and extensive communication that is a prerequisite for coordination and cooperation. 
Indeed, in order to have a smooth interaction, intention recognition is essential.
The DARPA/NSF final report on Human-Robot Interaction \cite{Burke2004} recommends to improve the models of human-robot relationship and in particular to work on the intentionality issue.

In \cite{Koo2009}, it is proposed to recognize intentional actions using relative movements of a human towards a robot.
An IR sensor embedded on the robot is monitored to track and estimate the velocity of a person.
They then infer intentional actions such as approaches and departs using Hidden Markov Models (HMM) and position dependent models.
\fif{Other related studies have defined human-robot proxemics in order to adjust inter-personal distance} \cite{walters2009empirical}.  
This work seems not enough to estimate engagement. 
\roy{On} one hand, one can slow down near the robot without wishing to interact.
\roy{On} the other hand, someone \roy{passing swiftly by close to} the robot might want to interact with it hastily.

The relative position and speed are not the only features that should be used to estimate intentionality.
\fif{Multimodal fusion and usage of postural information have given good results in the measurement of quality of human-robot interaction and engagement of the user into the interaction }\cite{castellano2009detecting, sanghvi2011automatic}. 
\fif{We aim to detect intention of interaction using also postural information and not only proxemics features.}

In his study, Knight \cite{Knight2011} points \roy{towards} the importance for a robot to convey and hence to detect intentionality.
It helps to clarify current activity and to anticipate goals.
Learning from the engagement of humans, the robot \roy{should} be able to anticipate the interaction and also to learn adequate moment when the robot itself can engage an interaction.
In \cite{Sidner2003}, engagement is defined as the process by which two (or more) participants establish, maintain and end their perceived connection during interactions they jointly undertake. 
Engagement is in the frame of connection that can be a collaborative task, spoken language, gestures etc. 
\fif{Sidner et al.} propose a model in three steps: (1) initiation of interaction, (2) \roy{sustainance} of interaction, (3) disengagement. 
As presented in \ref{sec:intro_theoapp}, we will see that our classification is based on this model.


\section{Corpus for Engagement with a Companion Robot}
\label{chap:env}
A part of the work accomplished was to build a multimodal dataset \roy{including} interactions with a companion robot equipped with a laser telemeter and \roy{a} Kinect device.
\roy{In this context, we focus on working} with consumer devices and in a natural and non intrusive manner. 
Even though the tendency \roy{is the increasing usage of} physiological sensors, such as R. Picard's pulse bracelet \roy{called} Cardiocam, physiological signals still \roy{remain an} invasive and \roy{relatively} expensive \roy{option} for users, \roy{for them to} be released widely.
The physiological modality is not considered in this work, yet it might be enriching to include it in \roy{a future} work \roy{that uses, for instance,} contact-free heart rate measurements \cite{Poh2010}.

In order to validate our hypothesis in the context of \roy{interactions with a} robot companion, the considered sensors are the ones commonly \roy{found} on such robots: microphones, video sensors, depth sensors \fif{and telemeters}.
There exist available datasets in the field of social signals processing dealing with non-verbal communication \roy{which use multiple} sensors.
These datasets for emotion recognition are unfortunately more often \roy{built} for face{-}to{-}face interaction where people sit and interact only with speech.
The SSPNet association \roy{released} the SEMAINE-DB dataset \cite{SEMAINEDB} where several persons \roy{were} recorded in a face-to-face speech interaction.
This database is suitable for a desktop environment \roy{that involves an} interaction with \roy{a} virtual commmunication agent.
It \roy{is not well suited for} human-robot interaction; especially as \fif{body} cues \roy{in} social signals are more diverse than facial expressions and speech characteristics.
\roy{There exist} other datasets using the Kinect sensor; such as Cam3D dataset centred on facial and hand movement associated with audio recording \cite{Mahmoud2011}, or the LIRIS  Human activities dataset \cite{Wolf2012} \roy{associated with} human activity monitoring task.
\roy{However}, the proposition of a robot centred dataset for multi-modal social signal processing has not been made yet.

Looking at limitations of the existing multimodal datasets, the sensors equipping the \kompai{} robot have been used to record a new robot view-point dataset where the \fif{users are interacting with the robot while standing}.
The scenarios included in this dataset will be presented \fif{in section} \ref{sec:senar}.

\subsection{Realistic Dataset}
\label{sec:acqu}

R. Picard in \cite{Picard05} states five variables that may affect data collection.
\fif{(1) The first factor is the spontaneity of the behavior. The emotion can either be elicited by a stimulus or acted.
(2) Another influence can come from the environment of the recording and the question here is: are the expressions of the participant similar in a lab setting and in a real-life situation? 
(3) Next question to be considered when recording affective data is: should the focus be on the expression or on the internal feelings of the participant?
(4) The participant's awareness about the fact that he's being recorded. Indeed, what is the influence of open-recording in comparison with hidden recording on the recorded data?
(5) Finally, should the participant be informed of the purpose of the experiment?

Regarding this research matter, (1) the engagement is relatively spontaneous, because the participants didn't act the interaction but were asked to interact whenever they wanted to with the robot. 
(2) The recording is made in a \textit{living lab} environment, similar to a flat.
The participants have no prior experience of this environment. This can create some fluctuations in their behavior. 
(3) We wanted to record intentionality of interaction, hence we focused on expression of social cues rather than to do a subjective evaluation.
(4-5)We chose to not tell the participants that we were interested in the social cues of intention of interaction in order to collect more natural data. 
}

\subsection{Experimental implementation}
\label{sec:interaction}
\begin{figure}[h!t]
\centering
 \includegraphics[width=0.4\linewidth]{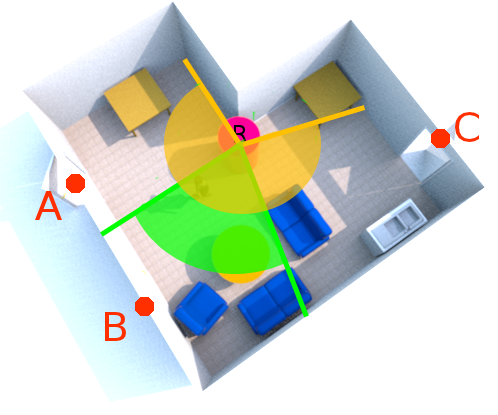}
	\caption{Home-like environment for our experimentation. The area has an L shape and 3 access doors (A, B and C). It is organized around 3 spaces: a living-room (near door B), a kitchen space (near C) an empty space (near A). The robot is place in the center of the area (purple cylinder). The view from the Kinect is depicted in green and the telemeter field of view in orange.}
	\label{fig:home_like}
\end{figure}

The experimentation space is presented in figure \ref{fig:home_like}.
The \roy{apartment is divided into} 3 \roy{areas}: a living-room, a kitchen space and an empty space.
To test our assumption \roy{that} spatial information is not enough to detect an intention of interaction, furniture is placed so that participants will need to pass near the robot each time they want to go from \roy{one} side to another of the experimentation room, even if they do not want to interact with \roy{the robot}.
This choice is an adverse condition as it leads us to distinguish someone passing close to the robot with or without intention of interaction.

In our recording, the robot is immobile.
All features are robot centered but can also be computed with a mobile robot.
The interaction in this dataset consist\fif{s} \roy{of playing} a ``tap~the~mole'' game on the mounted tablet PC on the robot.
Our hypothesis was that the interaction with a companion robot is also preceded by a pre-interaction phase (see section \ref{sec:intro_theoapp}) where the participant shows some subconscious social signals of its intention of interaction.
\fif{We also assume} that these cues are detectable with the sensor\fif{s} that \fif{are} equipped \roy{in} our enhanced version of the companion robot (see figure \ref{fig:kompaieq}).

\subsection{Steps of the interaction process}
\label{sec:intro_theoapp}
Sidner et al.  in \cite{Sidner2003} proposed a model \fif{to describe the} process of interaction in three steps: (1) initiation of interaction, (2) maintenance of interaction and (3) disengagement.
\fif{Our} work follows this approach by modeling these events as (illustrated in section \ref{sec:sample}): (1) \cwinter{}, (2) \cinter{},  (3) \clinter{}.
We added two more classes.
The \csomeone{} event is when someone is detected in the room but with no wish of interacting with the robot.
When nobody is in the room, it corresponds to the \cnoone{} event.

\subsection{Scenarios}
\label{sec:senar}
The data \roy{is} recorded \roy{in} two different scenarios performed several times by one or several participants in a home-like environment where the \kompai{} robot is \roy{present}. 
\doms{
The first one is dedicated to mono-user experiment.
Only one user will be in the room at a time.
The multi-users scenario addresses a more adverse condition.
Three persons are already in the room and interact with each other.
The idea behind this scenario is to check if we can detect starting engagement among social interaction between participant.
}

Each participant was given randomly one or several actions to perform in the room. 
As said, the room is similar to a small flat (Figure \ref{fig:home_like}). 
It was asked to the participant to enter the room by different doors, perform some realistic actions and \roy{to go} out.
One action is to interact with the robot.
\doms{
The other actions were going across the room, walking, sitting, playing cards or pouring water from the sink.
}

\begin{figure}
\centering
\begin{minipage}[t][.4\linewidth][t]{.45\linewidth}
 \centering
 \includegraphics[width=\linewidth]{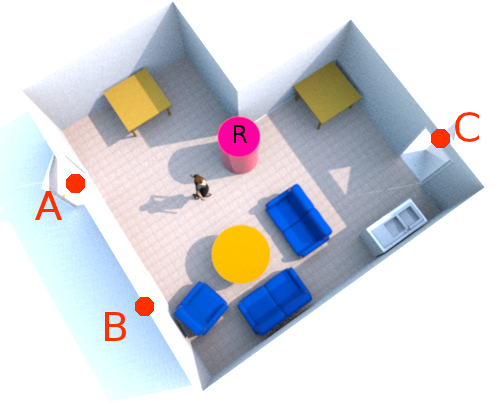}
 \caption{Scenario 1, Passing by}
 \label{fig:senar1}
\end{minipage}
\hspace{0.5cm}
\begin{minipage}[t][.4\linewidth][t]{.45\linewidth}
 \centering
	\includegraphics[width=\linewidth]{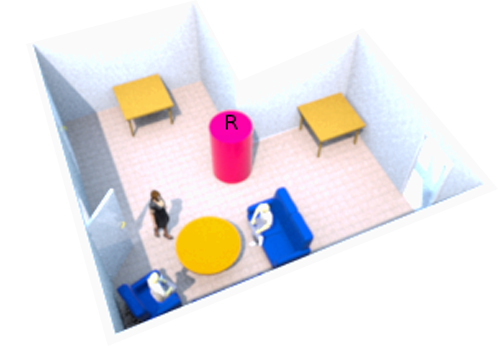}
	\caption{Scenario 2, Playing cards together}
	\label{fig:senar2}
\end{minipage}
\end{figure}

\subsubsection{Scenario 1: Passing By}
\label{senarios:1}
In this scenario, each participant \roy{is} asked to go through the room by different doors (A), (B) or (C).
\doms{
At this point, the given instructions did not mention the robot's presence in the living lab.
Participants were not aware that they will interact later with the robot.
}
After some crossings, the participant was invited to play the game on the robot's tablet.
The Figure \ref{fig:senar1} shows the setting of this scenario.

\subsubsection{Scenario 2: Playing cards together}
\label{senarios:2}
\doms{
In this second scenario, 3 or 4 persons were asked to enter the room and start playing cards in the living-room area.
}
A telephone placed in the room was used to ask one of the participants to execute an action (interacting with the robot, or using the sink for instance).
Once the participant was asked to perform a task, he could do it when he wanted to. 
The participants could sit wherever they wanted in the living-room area.
The figure \ref{fig:senar2} shows this scenario when a new participant is entering in the room while two participants are already sitting.

\section{Features Extraction}
\label{chap:theoritical-approach}
In order to characterize the engagement, features were extracted from the corpus previously introduced and then synchronized with a unique time scale.
Our \kompai{} robot, loaned by our partner Robosoft\footnote{\url{http://www.robosoft.com/}} (see figure \ref{fig:kompaieq}),
is composed of a mobile platform containing the wheel actuators, obstacle detection system, manual remote control utilities, etc.
The mobile platform is topped by a tablet serving as interface with the user, a pair of microphones, a motorized web camera and a speaker device.
We added a Kinect sensor to the robot.
Novelty of this work is not the adjunction of the RGB-D device but the synchronous usage of information from all sensors to compute a multimodal feature set.
The current version of the system is online and computes on the fly all features on the \kompai{}.

\begin{figure*}
\centering
\fbox{\includegraphics[width=0.98\textwidth]{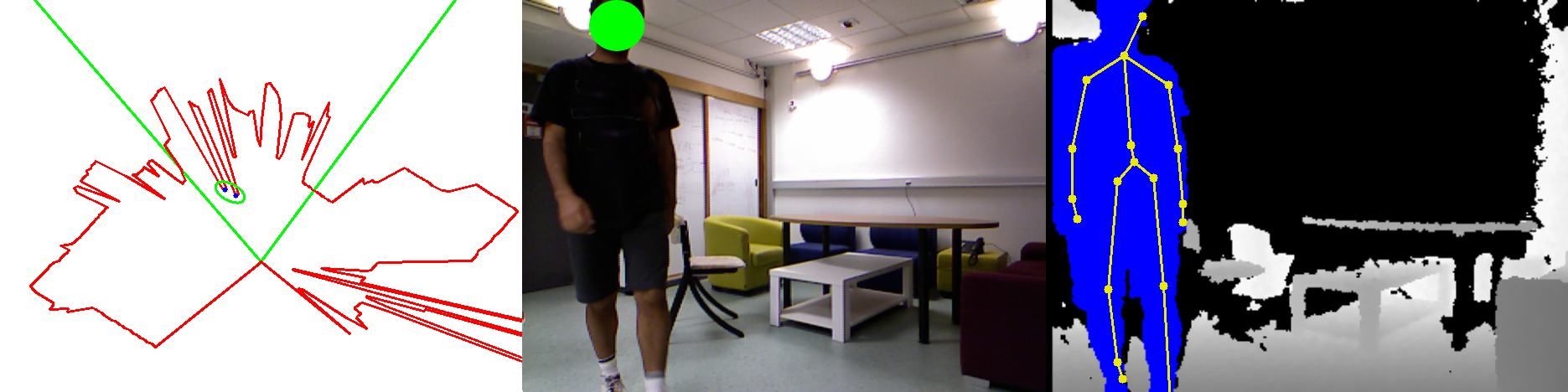}}
\caption{Perception from the robot point of view with a coming user toward a robot. Laser Telemeter (red lines), foot (blue spots) and pedestrian (green ellipse) information are depicted on left picture. In the middle, one can find RGB view from the Kinect with face detection (green circle). Right, the depth view with user (blue) and skeleton (yellow) tracking are drawn (note that with the first Kinect version, \textit{Kinect for XBox}, there is a little shift between the RGB and depth views). Acoustic and other body features are computed on these data (see section \ref{chap:theoritical-approach}).}
\label{fig:FullPerception}
\end{figure*}

Feature extraction algorithms \roy{present a fair amount of noise in general} and an interest of using multimodality is to be able to compensate one modality with another.
The full feature set gathered and computed on our corpus is composed of 99 features.
Then, a feature selection is made driven by social and cognitive science research on non-verbal communication cues depicted in section \ref{chap:related-work},
\doms{
the availability of sensors on our \kompai{} robot and the performance of algorithms in our experimental conditions.
We let aside important cues that might improve our results but are not usable in our context. For example, gaze direction and facial emotion recognition can not be computed; hand state and gestures are not reliable for instance.
Raw (\textit{x}, \textit{z}, \textit{y}, \textit{confidence}) tuples for skeleton joints permit us to compute more interesting body features.

This section presents different feature extraction techniques used in our experiment.
We chose to investigate a selection of 32 features including spatial information (spatial subset), body pose and video face detection (body subset), speech activity detection and sound localization (acoustic subset) in order to model the intention of engagement.
}
We detail these subsets in the following subsections.
These features are computed on several raw data channels: laser telemeter data coming from the robot, rgb video (section \ref{subsec:video}), depth view (section \ref{subsec:ske}) and audio channels from the Kinect (see figure \ref{fig:FullPerception}).
The synchronization and labeling methods are then explained in section \ref{sec:sync}.

\subsection{Spacial features}

\begin{figure}[!ht]
\centering
\fbox{\includegraphics[width=0.60\textwidth]{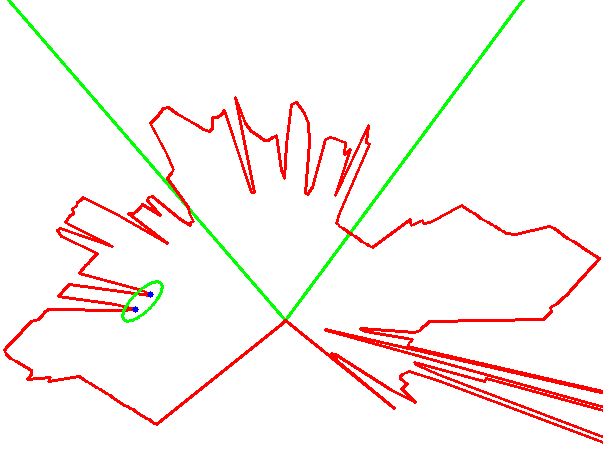}}
\caption{Spatial features: foot and pedestrians tracking using robot telemeter. Blues points are foot, ellipse represents tracked pedestrian.
The green lines represent, in the robot frame, the Kinect field of view. On this figure, there is one tracked pedestrian. The pedestrian is out of
the camera view.}
\label{fig:spatialfeatures}
\end{figure}

Proxemic features are classically used to describe role, attention and interaction, and in particular to determine the intention of interaction.
The tracking of the human trajectory can be done through visual based models or using laser telemeters.
Telemeters provide planar information of the environment \roy{while} covering a wider range angle than standard video camera with a good precision.
The \kompai{} robot is equipped with a single-row laser-range scanner at 20 cm above the ground.
For pedestrian tracking, it is more likely that we can detect shins.

Classical proxemic features are the relative position of the individual to the robot and his speed.
For a successful collaboration, the distance between the robot and the human should be optimum and the speed controlled \cite{Koo2009}.
It is important to know about distance so the person interacting with the robot does not feel uncomfortable \cite{hall1969}.
\cite{ZhaoS05} proposed a system for tracking pedestrians using multiple single row laser scanners.
Their pedestrian's walking model is described and used to accurately track pedestrian feet according to their swinging phase.
In our study, since a single scanner is used, foot occlusion is frequent as soon as a foot goes behind \roy{the other} from the telemeter point of view.
Therefore, it is difficult to predict swinging phases because of sparse data.
\doms{
Our human detection and tracking process is done through a feet-pairing process where a human is represented either by its both feet or one single foot when the other one is temporary hidden.
}

\subsubsection{Feet detection}

The laser sensors that equip the \kompai{} robot \roy{give} the distance values over 270 degrees \roy{every $80$ms}.
An adaptive background subtraction on the telemeters values is used to detect moving objects in the room.
Moving objects are candidates \roy{for being} feet.
The distance and the angle associated to the detected moving object are used to compute the positions $x$ and $y$ into the robot reference frame (see fig. \ref{fig:spatialfeatures}).

\begin{figure}
\centering
\begin{minipage}[t][.35\linewidth][t]{.45\linewidth}
 \centering
	\fbox{\includegraphics[width=0.8\textwidth]{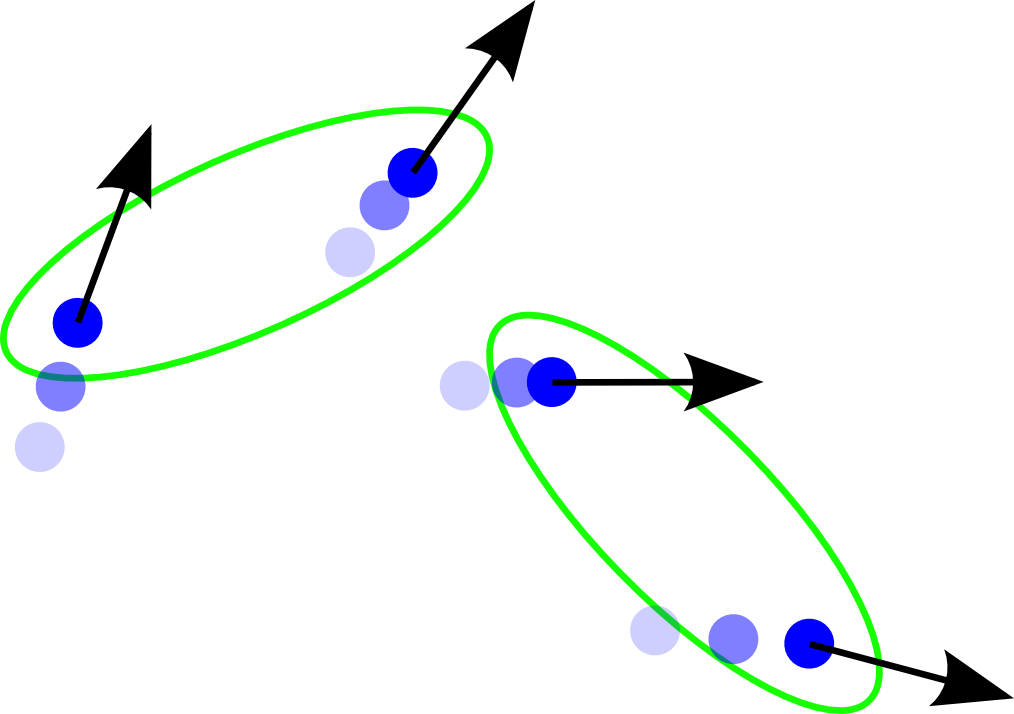}}
	\caption{Feet grouping if the pair of foot satisfies a constant space between legs through time.}
	\label{fig:pairing}
\end{minipage}
\hspace{0.5cm}
\begin{minipage}[t][.35\linewidth][t]{.45\linewidth}
 \centering
\fbox{\includegraphics[width=0.8\textwidth]{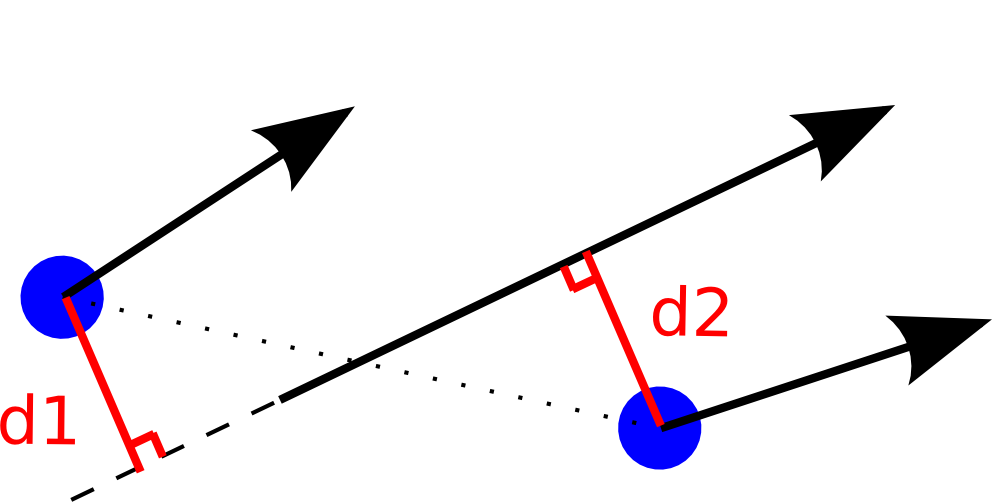}}
\caption{Space between legs is given by the sum of d1 and d2, the distances between feet and the main direction vector.}
\label{fig:legspace}
\end{minipage}
\end{figure}

\subsubsection{Speed Estimation}

A Kalman Filter is used on the set of moving objects detected by the laser to compute the speed and the acceleration at each frame.
The Kalman filter is an iterative prediction estimation algorithm allowing to introduce measured data (in this case the position $x$ and $y$ of a moving object) and to estimate dynamics such as the position and speed in two dimensions ($cible\_dx, cible\_dy$).
\doms{
The implementation of the Kalman Filter over the telemeters moving object data has been made using the OpenCV library.
}
A direction vector is extracted from each foot tracker.
It will be used for feet pairing in pedestrian tracking.

\subsubsection{Pedestrian tracking}

Grouping feet simply according to the distance between feet is not sufficient enough since pedestrians have different step length. \roy{Furthermore} people standing side by side could be misidentified.
Looking at the space between legs rather than between feet leads to a more robust parameter: even if the step length \roy{varies} for \roy{the} same pedestrian (standing, walking, running) the distance between legs along a direction vector remains constant during a natural walk because of the geometric properties of the human skeleton.
\doms{
Our process consists in pairing tracked feet that match a particular model, using a 2 stage filtering.
First, for each frame, feet that are less than one meter apart are paired up together forming a potential pedestrian.
This one meter threshold was empirically set to quickly exclude impossible pedestrians.
Then, candidates are evaluated and an actual pedestrian is revealed if it satisfies that the space between legs through a short frame sequence is relatively constant ({$\sim$}30~cm) as shown
}
in figure~\ref{fig:pairing}.

The space between legs is computed as the sum of the distance between each foot and their projection on the main direction vector, which is the sum of two single foot direction vectors, figure~\ref{fig:legspace}.
Pedestrian targets can be initialized as soon as they are in movement.
One of the feet can \roy{then} be hidden without causing the \roy{loss
of padestrian's localisation}, it will be paired up when the foot appears again.
If both feet disappear, then the pedestrian tracker is lost and deleted.

Our tracking process is capable of tracking multiple walking pedestrians with frequent occluded feet from a single range laser telemeter.

\floatstyle{plain}
\restylefloat{table}

\begin{table}[h]
\centering
\begin{tabular}{c c c }
\textbf{Features Name} & \textbf{Sensor}  & \textbf{Frequency}\\
\hline \\ [-1.5ex]
Positions $(cible\_x,cible\_y)$ & laser range finder & 12.5Hz \\ \hline \\ [-1.5ex]
Speed $(cible\_dx, cible\_dy)$ & laser range finder &12.5Hz \\ \hline \\ [-1.5ex]
Distance ($cible\_dist)$ & laser range finder &12.5Hz
\end{tabular}
\label{tab:space}
\caption{Features from the Space subset}
\end{table}

Every 80~ms, we have the \textit{Number of pedestrians} around the companion robot and for each pedestrian, we get an \textit{id}, \textit{cible\_x} and \textit{cible\_y} position in the reference frame of the robot, his distance to the robot \textit{cible\_dist} and \textit{cible\_dx}, \textit{cible\_dy} the speed of the pedestrian in \textit{x} and \textit{y} axis (see Table  1 above).

\subsection{Acoustic features}
\label{subsec:audio}

Pantic in \cite{Pantic2003} and \cite{Vinciarelli1920} list some features from the audio signal that can be used to spot basic emotions such as happiness, anger, fear and sadness.
It can be agreed on that some audio features such as pitch, intensity, speech rate, pitch contours, voice quality and silence are good parameters to classify the emotional state of an individual.
Moreover, speech is an important information source for social glue with a companion robot \cite{Auberge2013}.
Considering the recognition of the starting engagement in an interaction, only few papers in the literature use audio features in a multimodal frame.
\cite{Ooko} proposes an engagement estimator using head pose associated to audio features in a face-to face conversational agent sitting interaction.
Some articles invoke interest of sound localization in attention or focus estimation \cite{Maisonnasse2006}.

\begin{table}[h]
\centering
\begin{tabular}{p{5cm} p{4cm} l}
\textbf{Features Name} & \textbf{Sensor}  & \textbf{Frequency}\\
\hline \\ [-1.5ex]
Speech Activity $(sad\_event)$ & Kinect's Microphones & 100Hz \\  \hline \\ [-1.5ex]
Source\hspace{0.2cm}localization $(beam, angle, confidence)$ & Kinect's Microphones & 8Hz
\end{tabular}
\label{tab:acou}
\caption{Features from the Acoustic subset}
\end{table}

The microphone array embedded in the Kinect sensor is a four-element linear microphone array processing acoustic echo cancellation and noise suppression.
Using this audio stream, we can compute Speech Activity Detection
(SAD) \cite{Vaufreydaz2006}, which \roy{is indicative of the
  parts} of \roy{the} acoustic signal \roy{representing} speech.
The SAD labels the audio stream every 10~ms.
The source localization outputs the stimulated beam (rough estimation) and the source position (more accurate angle) associated with a confidence.
The frame rate of the acoustic localizer is 8Hz.

\subsection{Body features}
\label{subsec:ske}

The Skeleton tracking of the Kinect sensor allows real time pose and gesture recognition.
Our system outputs at depth camera frame rate the \textit{number of skeleton}, and for each skeleton an \textit{id} and 60 features giving \textit{x}, \textit{z}, \textit{y} and \textit{confidence} for each joint.

\begin{figure}[h!t]
   \centering
   \fbox{\includegraphics[width=0.45\linewidth]{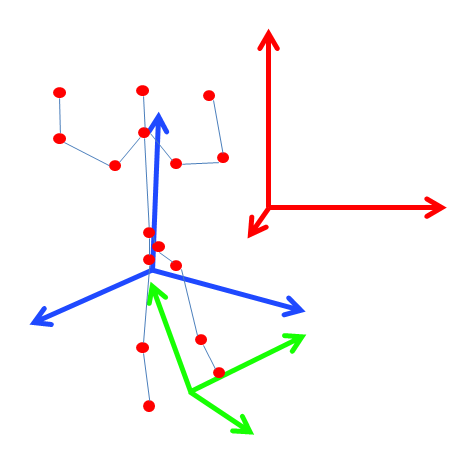}}
 \caption{Stance (green) pose, hip (blue) pose and torque. Body pose features computed from the skeleton information of the Kinect sensor (positioned in red)}
 \label{fig:BP}
\end{figure}

\subsubsection{Body pose}
\label{sec:bodypose}

As expressed previously, body pose features give clues on intention of interaction.
These features can measure the level of engagement of a user into a task as in \cite{Sanghvi}, proposing a measure of the Body Lean Angle.
Psychologists have proposed many models to describe body pose metrics and their associated meaning.
An overview of these metrics can be found in \cite{Mead2011}.
In \cite{Holthaus2011}, the authors propose a spacial model coupling attention estimation and distance metrics for a receptionist robot to infer the intentions of the human.
Their results are promising, but the approach is limited to face-to-face interaction.
\cite{huttenrauch2006} studied on spatial relationships in human robot interaction and concluded that human-human proxemic measures and social arrangement such as Hall's interpersonal distance system are not enough to achieve socially appropriate robot behavior.
Psychologists such as Hall \cite{hall1969}, Mehrabian \cite{Mehrabian1996} Schegloff \cite{Schegloff1998} have proposed some metrics that have been used in computer assisted analysis of posture, but there is no consensus on one particular model.
Posture is difficult to measure and evaluate using computer vision.
Nevertheless, with new devices like the Kinect Sensor and other real-time 3D pose reconstruction systems, we are now able to evaluate the pose of a person.

The body features used in our experiments are based on Schegloff metrics presented in \cite{Mead2011,Schegloff1998} and computed from the Kinect skeletons.
These features aim to depict the body pose of the individual.
The accent is posed on the stance, the hips, the torso and the shoulders' position and orientation relatively to each other.
We obtain 19 features containing Schegloff's metrics at depth frame rate.

\subsubsection{Distance}
What is interesting about body features is that they depict the orientation of the bodypart relatively to the Kinect sensor placed on the robot.
A \textit{skeleton distance} associated to the skeleton position is computed using the average z-value of several joints of the skeleton.

\subsubsection{Face detection}
\label{subsec:video}

In terms of affect \& emotion detection and speech recognition, a lot of studies have published results with a combination of face and audio features \cite{Rich2010a, Silva2004a, Chen1998, Karpouzis}.
Within engagement, the orientation of the head and the gaze seem to be crucial.
As shown in \cite{Rehg1999}, a speaker can be detected more easily with the combination of different features as a mouth sensor.
Face detection is already a first cue of interaction, and orientation of the face toward the robot is a reliable sign of attention.
Gaze tracking can give a better estimation of user' attention, but performance in uncontrolled real-life condition (small faces in video stream or untrained gaze angle for instance) are not good enough.

From the video extracted from the RGB stream, we propose to focus on face detection.
We \roy{use} a trained machine learning system using Haarcascasdes method.
The training is provided by the OpenCV library \cite{Haarcascade}.
The gathered  features are the position of face(s) in the pixel reference frame, the \textit{\{0,0\}} point is the center of the image.
For each detected face, we have \textit{x}, \textit{y} and \textit{face size}.

\begin{table}[h]
\centering
\begin{tabular}{p{7.5cm} p{2.8cm} p{2cm}}
\textbf{Features Name\footnotemark} & \textbf{Sensor}  & \textbf{Frequency}\\
\hline \\ [-1.5ex]
Stance $(*Pose\_x,*Pose\_y, *Pose\_z,$ $*Pose\_rot)$ \hspace{0.5cm}for feet, hips, torso and shoulders & Kinect's Skeleton & 30Hz \\ \hline \\ [-1.5ex]
Relative torque angle $(*Torque)$ for hips, torso, shoulders  & Kinect's Skeleton & 30Hz \\ \hline \\ [-1.5ex]
Skeleton distance $(skl\_dist)$ & Kinect's Skeleton & 30Hz \\ \hline \\ [-1.5ex]
Face $(face\_x, face\_y, face\_size)$ &  RGB stream & 30Hz \\
\end{tabular}
\label{tab:bodyp}
\caption{Features from the Body pose subset.}
\end{table}
\footnotetext{Suffixes are presented with a ``$*$'' character. For example, we compute \textit{shoulderPose\_x}, \textit{shoulderPose\_y}, \textit{shoulderPose\_z}, \textit{shoulderPose\_rot} and \textit{shoulderTorque}.}

\subsection{Fusion, Synchronization and labeling of features}
\label{sec:sync}

\doms{
At this point, we have a selection of 32 features: pedestrian
information (\textit{x}, \textit{y}, \textit{speed\_x},
\textit{speed\_y} and \textit{distance} to robot), Schegloff metrics
(computed from the skeleton
}
see \ref{sec:bodypose}), face detection, speech activity detection and sound localization.
\doms{
This section details how we deal with sparse features, features synchronization and corpus labeling.
}

\subsubsection{Sparse features}
\label{sec:sparsedata}

Multimodality has one major drawback.
Space coverage is not the same for all sensors: the Kinect has a 60 degrees field of view, the laser telemeter 270 degrees, etc.
Moreover, we do not have every feature all the time.
Whereas video, depth and laser telemeter data, face detection, sound classification, skeleton and pedestrian tracking are not available all the time.
One way to cope with these sparse data is to train several classifiers with all possible combinations for available features and to select the adequate one at the right moment.
The problem with this approach lies in reducing the amount of data for training for each subtype of classifiers.
Another way to solve this problem is to use specific neutral values for unavailable features.
For example, when there is no pedestrian, we can set all pedestrian features (position, speed and acceleration) to 0.
This set of values is considered neutral as it is impossible to find them in observed data.
In these experiments, as we \roy{did} not have enough data to train each subtype of classifiers, we chose the second method.

\subsubsection{Features synchronization}

We needed to synchronize the monitored data from the different modalities.
Data collected through the Kinect sensor such as the skeletons{'} positions, the video and the depth are tagged with a time relative to the Kinect sensor{'}s initialization.
The laser data are labeled with an absolute time stamp thanks to the real-time micro-controller of the \kompai{} robot.
The telemeters' input is the steadiest one at a fixed 80~ms period, hence it is used as synchronization frame rate at 12.5HZ.
The short time delay between frames prevents us to interpolate and allows to elicit the last value of each feature as the current value.

\subsubsection{Corpus labeling}

The labeling of the dataset with the 5 classes (\cwinter{}, \cinter{}, \clinter{}, \cnoone{}, \csomeone{}) is semi-automatic.
\doms{
The timestamped notes of the experimenter contain start time and end time of all events (participant \textit{p} is entering the room, a specific action was asked, etc.).
This annotation serves as first segmentation input.
The first labels are then made automatically using both the tablet touching information and the available features.
}
The \cinter{} time-interval is labeled from first touch of the tablet until the last click.
A \cwinter{} event starts when someone is coming to the \kompai{} robot before an \cinter{} event.
If someone is moving or sitting, and after decides to come to the robot, only the direct path to the robot is labeled as \cwinter{}.
The \clinter{} labeling is made like the \cwinter{} even and corresponds to all direct paths after leaving the interaction.
The \cnoone{} and \csomeone{} events, correspond to the rest of the time, respectively, there is no one in the room and when a person is present, but there will be no interaction.

All labels have been reviewed by a human expert looking at the video recordings.
\doms{
No problems were reported as a result.
}

\subsection{The dataset in numbers}
\label{subsec:datanb}
\begin{figure}[H]
\centering
\includegraphics[width=0.75\linewidth]{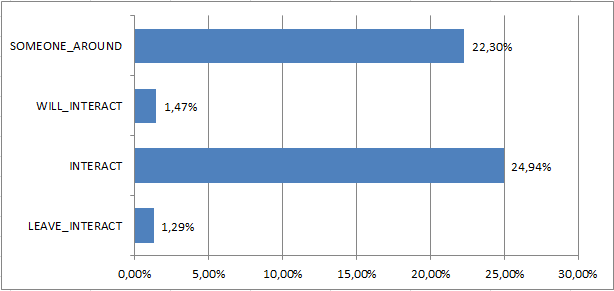}
\caption{Percentage of each class in the dataset.}
\label{lab:chart}
\end{figure}

\doms{
In our dataset, each frame corresponds to 80 ms, provides a full feature set and has a unique label.
The total number of recorded frames is 158200\footnote{The total recording time is 3:30:56.}.
The number of synchronous frames for each event is not equal.
}
The Figure \ref{lab:chart} shows the data distribution of each event.

\doms{
In real life, individuals do not express social signals the same way.
A certain variability was introduced in the pool of 19 participants.
They were from 20 to 35 years old, almost 50\% male/50\% female, students, administrative assistants and researchers.
Voices, clothing styles (colors, trousers or skirts, etc.) and statures vary to challenge perception algorithms.
15 participants did 1 or 2 interactions from 2 to 10 minutes according to their will, 9 were recorded both in mono-user and in multi-users scenarios.
}

\doms{In total, the corpus includes 29 interactions with the robot, made by 15 different participants.
}
The total size of the uncompressed data set is around 300~GB.
One can find samples of the corpus in \ref{sec:sample}.

\section{Multimodal detection of engagement}
\label{chap:evaluation}

\fif{In this work, we first choose }to test all the modalities that can help us to detect intention of interaction. 
Then, a selection can be made among the most relevant multi-modal features (section \ref{sec:MRMR}). 
\doms
{
The evaluation focuses on comparing the detection of the intention of interaction by using multimodality versus simple spatial information.
We want to confirm that these state-of-the-art approaches based on spatial features only are not enough in home-environment with furniture.
}
The Scipy library through Sklearn \cite{SVM} and the Weka toolbox \cite{Weka} were used for the classification.
The techniques used for the classification are the Multi-class Support Vector Machine (SVM) from Sklearn and the Artificial Neural Network (ANN) technique from Weka.

\subsection{Prepare the dataset for the Classification (K-Cross Folding)}
\label{sec:prep}

In order to train a model and to test it afterwards, the dataset needs to be split in a training set and a test set.
A way to randomize this splitting is the k-cross folding process. 
In this method, the dataset is partitioned in $k$ subset. One subset is kept for testing and the $k-1$ others are used for training the model.
This splitting process is repeated $k$ times so that each subset is used once for testing against others subsets.
K-cross validation allows to ensure that the splitting is quite random.
Since the events (interaction phases) are not equally probable and temporally related, we used a \textit{stratified} k-fold-cross validation that keeps the same proportion of the different classes in the splitting process.

\doms{
For our experiment, using $k=10$, the train and test sets are composed respectively of 140292 and 15587 frames\footnote{Note that the \textit{stratified} splitting process let aside 2321 frames.}.
}

\subsection{First classification experiment}
\label{sec:class1}

We chose to use two kinds of classification in this experiment.
Even if many other techniques could have been applied, we decided to focus on Neural Network and Support Vector Machine (SVM) (sections \ref{ssec:ann} and \ref{ssec:svm}). 
For this two techniques, we built and tested two classifiers one for the multimodal condition (including 32 features) and one for laser telemeter only condition (a subset of the multimodal dataset including spatial information only), see \cite{benkaouar:hal-00735150}. In \ref{sec:MRMR}, we try to determinate if some features are more relevant for our task.

\subsubsection{Neural Network}
\label{ssec:ann}

\floatstyle{plain}
\restylefloat{table}

\begin{table}
\centering
\begin{minipage}[t][.20\linewidth][t]{.43\linewidth}
\resizebox{\linewidth}{!}{
\begin{centering}
 \begin{tabular}{c c c }
 \multicolumn{3}{c}{\textbf{Telemeters condition}} \\ \hline
 Class	&Precision	&Recall\\ \hline
\cnoone{}	&0,95	&1,00\\
\cellcolor{yellow}\cwinter{}	& \cellcolor{yellow}0,91 & \cellcolor{yellow}	0,77 \\
\cinter{}	&0,77	&0,96\\
\clinter{}	&0,00	&0,00\\
\csomeone{}	&0,75&	0,35\\ \hline
\end{tabular}
\end{centering}
}
\end{minipage}
\hfill
\begin{minipage}[t][.20\linewidth][t]{.43\linewidth}
 \resizebox{\linewidth}{!}{
 \begin{centering}
 \begin{tabular}{c c c }
 \multicolumn{3}{c}{\textbf{Multimodal condition}} \\ \hline
Class &	Precision	&Recall\\ \hline
\cnoone{}&	0,95	&1,00	\\
\cellcolor{yellow}\cwinter{}	& \cellcolor{yellow}0,90	&\cellcolor{yellow}0,87 \\
\cinter{}	&0,84&	0,95 \\
\clinter{}	&0,21&	0,01 \\
\csomeone{}	&0,76&	0,41 \\ \hline
\end{tabular}
\end{centering}
}
\end{minipage}
\caption{Results for Neural-Network 5-classes classification using Weka. Left table presents results for the telemeter condition. Right table for the multimodal condition.}
\label{tab:resNN}
\end{table}

The Artificial Neural Network is a graphical layered model commonly used to infer model from observation.
In our case, we suppose that our features set can characterize the starting of engagement.
ANN is a good classifier to build prospective detection especially with large features vector. 
The test results of the ANN classification are presented in left table in \ref{tab:resNN} for the telemeter, and the right table for the multimodal dataset.
Notably, one can see that that \clinter{} was not classified in the telemeter condition.
The \cinter{} precision increased in multimodal condition combined with a small loss in recall.
Concerning the \cwinter{} class, the system returns more relevant events using multimodality (higher recall score) even if its precision decreased.
In multimodal condition, the precision is improved for most of classes.
The Neural Network classifier gives always better recall rate in this condition.

For the intention of engagement detection, in a practical point of view, the accent has to be put on the good performance in term of recall associated to a low false-positive rate.
Using Neural Network in multimodal condition seems to fulfill this requirement.

\subsubsection{Multi-Class Support Vector Machine}
\label{ssec:svm}
The results of the 5-classes classification using support vector machine for the multimodal condition are presented on right table in \ref{tab:resSVM}, on left table for the telemeter condition.
Analyzing these tables, one can see that the False-Positive rate is higher in the telemeter condition.
The \cinter{} class is not classified at all.
The precision and recall scores for \cwinter{} class are improved by the multimodality.
The aim of our work was especially to decrease the rate of misclassifying an event as \cwinter{}, hence the system has fewer chances to predict an interaction when there is one user with no intention of interaction.
In the case of an SVM classifier, multimodality is thus interesting for this purpose.

Scores for \cinter{}, \clinter{} and \csomeone{} classes are of interest.
In multimodal condition, the \csomeone{} scores drop while \cinter{} and \clinter{} are better classified.
This fact is a first clue of the closeness of our classes in the feature space.
The section \ref{sec:disc1} will discuss this topic.

\begin{table}
\centering
\begin{minipage}[t][.20\linewidth][t]{.43\linewidth}
\resizebox{\linewidth}{!}{ 
\begin{centering}
 \begin{tabular}{c  c c }
 \multicolumn{3}{c}{\textbf{Telemeters condition}} \\ \hline
Class &	Precision	&Recall \\ \hline
\cnoone{}	&0,68	&1,00	 \\
\cellcolor{yellow}\cwinter{}	&\cellcolor{yellow}0,80	&\cellcolor{yellow}0,68	 \\
\cinter{}	&0,00	&0,00  \\
\clinter{}	& 0,00 &	0,00	 \\ 
\csomeone{}	&0,76	&0,01	\\ \hline
\end{tabular}
\end{centering}
}
\end{minipage}
\hfill
\begin{minipage}[t][.20\linewidth][t]{.43\linewidth}
 \resizebox{\linewidth}{!}{
\begin{centering} 
 \begin{tabular}{c  c c }
 \multicolumn{3}{c}{\textbf{Multimodal condition}} \\ \hline
 Class	&Precision	&Recall	\\ \hline
\cnoone{}	&0,92	&0,88 \\
\cellcolor{yellow}\cwinter{}	& \cellcolor{yellow}0,92	& \cellcolor{yellow}0,71 \\
\cinter{}	&0,54	&0,77 \\
\clinter{}	&0,04	&0,10\\
\csomeone{}d	&0,52	&0,29\\ \hline
\end{tabular}
\end{centering}
}
\end{minipage}
\caption{Results for SVM 5-classes classification using Sklearn. Left table presents results for the telemeter condition. Right table for the multimodal condition.}
\label{tab:resSVM}
\end{table}

\subsubsection{Minimum Redundancy Maximum Relevance experiment}
\label{sec:MRMR}

A dimensional reduction of the features space was made using the Principal Component Analysis (PCA) and the Linear Discriminant Analysis (LDA) using the Sklearn tool-kit.
The results were not conclusive, the dimensionality reduction gave strictly the same performance during the classification where we were expecting an improvement.

The Minimum Redundancy Maximum Relevance \cite{MRMR05} (MRMR) technique was performed in order to highlight the best features for our detection system.
This dimensionality reduction technique has the advantage of giving the more relevant features instead of building new features \fif{from} the observed ones.
\doms{
Using mutual information, correlation and t-test/F-test metrics, the MRMR algorithm selects a feature subset maximizing dissimilarity of features and statistical characterization of the classification.
}
It could allow eventually to discard less relevant features in order to optimize the detection of engagement process.

\begin{figure}
   \centering
   \includegraphics[width=0.75\linewidth]{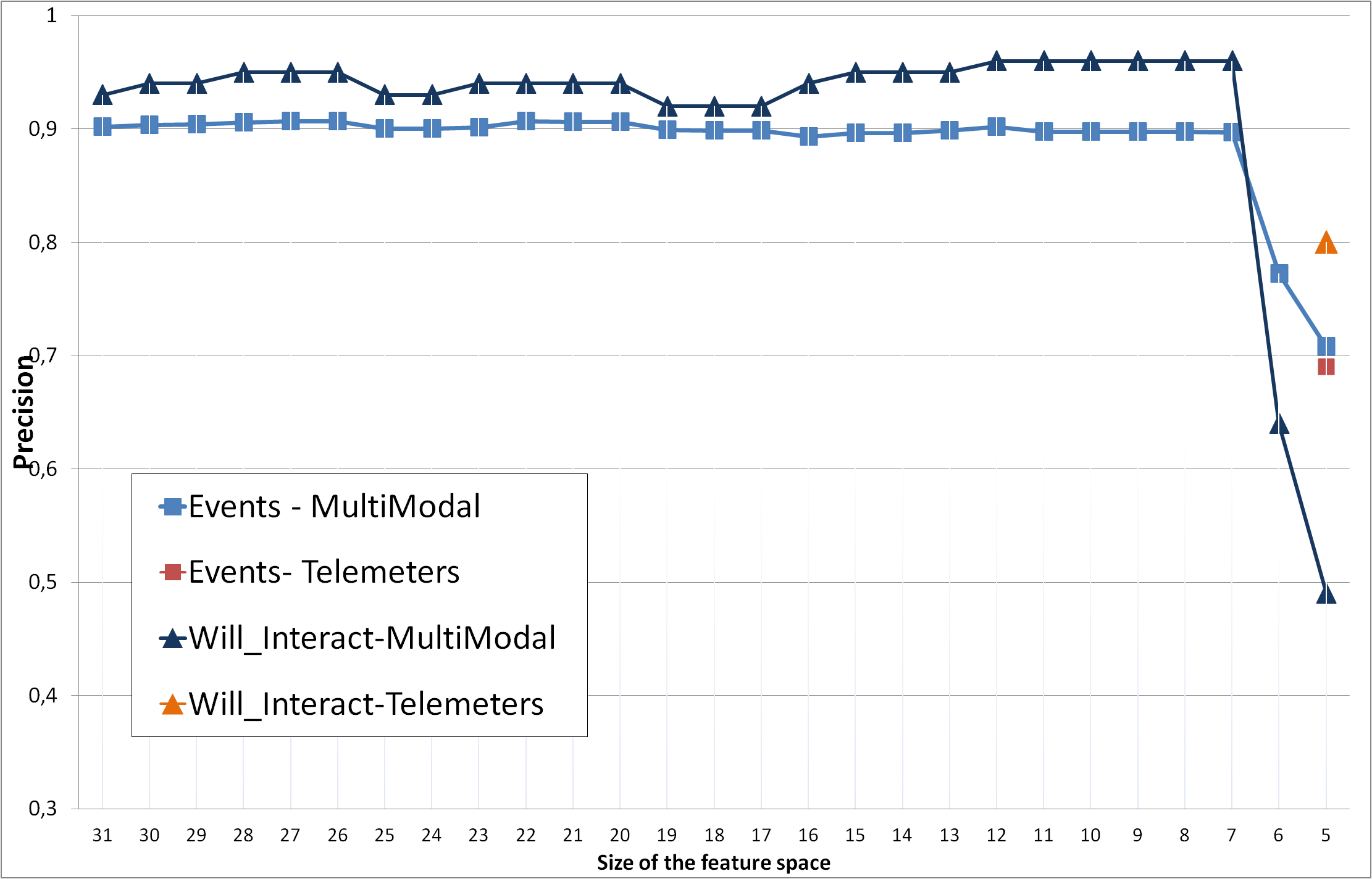}
\caption{Precision evolution with the decreasing number of multimodal features in comparison with the telemeter condition for all events and for the \cwinter{} event}
 \label{fig:mrmrpre} 
\end{figure}

\begin{figure}
   \centering
   \includegraphics[width=0.75\linewidth]{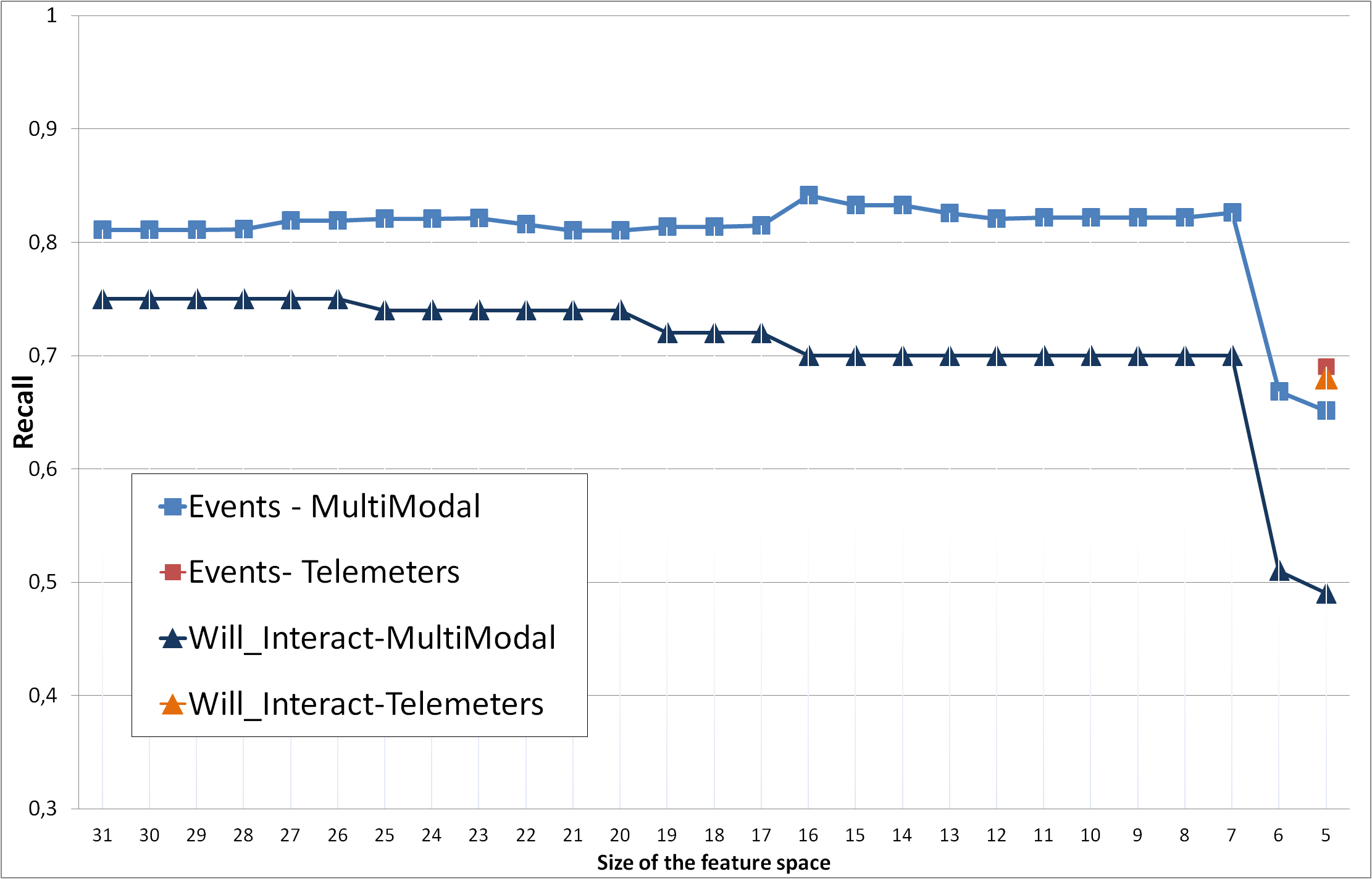}
\caption{Recall evolution with the decreasing number of multimodal features in comparison with the telemeter condition for all events and for the \cwinter{} event}
 \label{fig:mrmrrec} 
\end{figure}

In order to evaluate the relevant features for the multimodal detection of intention of interaction, we used a MRMR dimensionality reduction from a vector of 32 features before performing an SVM learning.
The Figure \ref{fig:mrmrpre} shows the feature reduction's impact on the precision.
The precision drops at the 6-features reduction. 
\fif{From the 32-features till 7-features along the feature space reduction, the precision remains pretty stable.}
These results confirm that there are many \fif{redundancies} in the 32-feature space.
Some of these features seems to be fundamental for a better detection with a higher precision than the telemeters' one.
Equivalent conclusions can be made on the Figure \ref{fig:mrmrrec} regarding the recall performances.

The first remark on these results is that the seven highest rated features are coming from heterogeneous modalities.
The \textit{face~size} and \textit{face~x} are respectively the relative size and position of the face in the Kinect view.
The \textit{beam} and the \textit{angle} are the sound localization features from the microphone array.
The telemeter information are considered as relevant, with the high selection rate of the speed \textit{speed\_x} and \textit{y} position.
The fact that these spatial features are selected among the most relevant ones in our multimodal set is not surprising.
They were part of previous state-of-the-art researches.
Last, the \textit{shoulder pose rotation} corresponds to the relative orientation of the shoulder in the body, and is extracted from the skeleton information.

\subsubsection{Discussion about first experiment}
\label{sec:disc1}

The previously presented results support our assumptions about multimodal recognition of intention of interaction.
The body pose, especially the  shoulders orientation were shown to be relevant for intentionality detection.
Spatial information coming from the audio and telemeter streams are also important.
The position and size of detected faces in the video confirms that facing the robot is a sign of intentionality of engagement.

These evaluations were conducted using an a priori selection of 32 features.
This selection was inspired by our literature searches in human-robot interaction, social sciences and cognitive science fields.
Results are improved, but can we conclude that our results are generic enough?
For instance, we replaced all skeleton information by Schegloff metrics.
Even if results validate our hypothesis, we need to check if we do not have interesting information in the left aside 67 other features.

From the results\footnote{As we did a \textit{stratified} k-fold-cross validation, we have many confusion matrices. Presenting one will not correspond to the k-fold-cross validation result (table \ref{tab:resNN} and \ref{tab:resSVM}), showing all is not possible.}, we see that \clinter{} is never well classified in the telemeter condition.
In the multimodal condition, the precision and recall scores get slightly improved.
Anyway, \clinter{} is most of time confused with \csomeone{}.
Several explanations may enlighten this result.
When someone is interacting with the \kompai{}, he is close to the robot.
We have spatial features computed from laser telemeter but no information about his body from the Kinect (see table \ref{tab:kinectv}).
A more intuitive point can be that less social signals are expressed when leaving interaction.
For closeness reasons, the \cinter{} class presents also low classification results.
We do not actually need to detect it: intention of engagement is a prior state to interaction.

In preliminary conclusion, we can say that our primary hypothesis is validated: multimodality can improve engagement detection on a companion robot with embedded sensors.
Nevertheless, other experiments must be conducted with a three classes approach (\cnoone{}, \csomeone{} and \cwinter{}) and all the available features.

\subsection{Second experiment}
\label{sec:class2}

In this second experiment, we will tackle our classification task in regards to the lessons learned in \ref{sec:disc1}.

\subsubsection{New 3 labels corpus}
\label{subsec:3labels}
\paragraph{Validation of our 3 classes}
We need to validate our hypothesis about the confusion of the \clinter{} and \csomeone{} classes.
We conducted clustering experiments using the k-means algorithm.
We wanted to check if it is difficult to separate these 2 classes in the features space.
K-means is an algorithm that produces the best clustering knowing the number of wanted clusters.
We ran K-means from 2 to 1500 clusters with our 7{-}features set and checked the distribution of each feature vectors in these clusters.
In all clusterings done, there is no significant separation between our 2 classes, i.e. the feature vectors of each class are equitably distributed among clusters.
We did clustering with every feature set up to our 32 features and distribution remained diffused.
This result also corroborates that it is likely to say that either people do not express strong social signals when they leave interaction with the robot or that in our hardware setup, we cannot compute them.

\paragraph{New labeling}

\begin{figure}[h!t]
\centering
\includegraphics[width=0.75\linewidth]{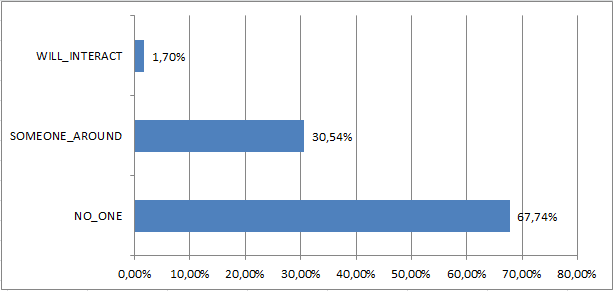}
\caption{Percentage of frames per event with 3{-}classes labeling}
\label{chart2}
\end{figure}

We modified our labeling using this time only our 3 classes (see discussion \ref{sec:disc1}).
All frames from the \cinter{} class were removed and all instances of \clinter{} were replaced by \csomeone{}.
From the remaining 124282 frames, using k-cross folding (see \ref{sec:prep}) with $k=10$, we computed train and test sets respectively of 111854 and 12428 frames.
Repartition of each class is given in the figure \ref{chart2}.

\subsubsection{Classification}

\begin{table}
\caption{Results of multimodal SVM 5-class classification using Sklearn with our 3 classes.}
\label{tab:resSVMMM2}
\centering
\resizebox{0.43\linewidth}{!}{%
\begin{tabular}{c  c c }
\\ \hline
Class	&Precision	&Recall	\\ \hline
\cellcolor{yellow}\cwinter{}	&\cellcolor{yellow}0,95	&\cellcolor{yellow}0,93\\
\cnoone{}	&0,93	&1,00 \\
\csomeone{}	&0.99	&0,84 \\
\end{tabular}%
}
\end{table}

Using the new dataset, we re-performed the classification process using SVM.
Results are shown in the table \ref{tab:resSVMMM2}.
We can see that we have an overall improvement in our results on the 3 remaining classes, mainly on the recall score.
We increased both scores on the \csomeone{} class.

Training on this data was more successful, but new experimentation in other conditions, in other places, with different lighting conditions, with more participants has to be done in order to conclude definitely on these results.

\subsubsection{Feature selection among all available features}

We re-ran our experiment using the MRMR technique to select in our total 99 features set the most relevant ones.
Results in this case differ from the first experiment.
Some intuitive features, like the \textit{facing coefficient} and the \textit{skeleton distance}, were selected.
Surprising features appear among the more relevant ones.
Indeed, the most important feature is the \textit{right ankle x} position.
This fact is peculiar when one knows that more than 70\% of the frames have no information about a skeleton.
Moreover, many skeletons have noisy feet information due to the position of the robot in the living lab.
During our recordings, participants passed very closed to the robot, as we wanted them to do (see \csomeone{} example in \ref{sec:sample}).
In this case, the mounted Kinect did not manage to compute confident 3D feet position. 

The MRMR technique is not efficient on this task.
For a 99 features space, we may not have enough data to determine reliable metrics (mutual information, correlation, t-test/F-test...) used by the algorithm.
\doms{
Nevertheless, this experiment confirms that selecting human readable features inspired from social and cognitive sciences could be an alternative method for feature space reduction.
}

\section{Conclusion}
\label{chap:conclusion}

Psychologists working on the acceptance of \roy{a} robot by \roy{the} elderly
and people with disabilities have pointed \roy{out} the need for more natural and acceptable interactions with the companion robot \roy{in the home environment}.
\roy{A} starting engagement is the first step of interaction.
It corresponds to the phase preceding the \roy{actual} interaction, when the user \doms{implicitly} \roy{signifies} his wish to interact.
For a companion robot, the skill to detect engagement \roy{, and thus,} to anticipate human will is a key feature to make it socially acceptable.

Our goal was to evaluate and measure \roy{humans'} cues for engagement \roy{} in an interaction with a robot.
\roy{They were} classically detected using the position and the speed of the user\roy{.} We have shown the limits\roy{,} in term of recall\roy{, in} performance of this technique\roy{,} confronting it \roy{with} realistic scenarios \roy{of} engagement towards a robot in a home environment.
Indeed, the proximity of the user with the companion robot is not a sufficient criteria \roy{when predicting} the engagement.
Spatial based detection of intention of interaction used in previous approaches gave good results in lab environment.
However, the congestion of home-environment leads to situations where humans pass close to the robot without the will of interaction and where spatial based detection gives false positive responses.

\roy{Having built} realistic scenarios involving the interaction of participants with the \kompai{} robot, we have collected \roy{sensory data of various} engagement \roy{sequences}.
Several features were computed over multiple modalities.
From the video, we detected the size and position of the face in the image.
The skeleton data gave us clues to compute body \roy{poses}.
The \roy{audio} was used for the sound localization and for the speech activity \roy{recognition}.
Telemeters gave us an estimation of the position and the speed of \roy{the} pedestrians.

A cross-fold validation allowed us to segment our dataset into training and testing sets.
These subsets where used by two different classifiers, a Neural Networks and a Support Vector Machine.
These classifiers\roy{,} trained on multimodal and telemeters features set\roy{,} gave better performances for the multimodal condition.
This fact showed that spatial and speed features used in related works are not enough in a home environment.
Multimodality improved the recall of the engagement detection \roy{significantly}, \roy{which} was the \roy{}hypothesis of this research.

\subsection{Key points}

\doms
{
Transposing social and cognitive sciences results and using human readable features can be an alternative approach for feature selection.
Using this methodology, we enhanced spatial information with the selected body related and acoustic features and get better detection scores, notably in term\roy{s} of recall.
As far as we know, we validated experimentally for the first time, that shoulder pose rotation, a metric from Schegloff's research in Sociology, is of importance to the detection of intention of engagement.
}

The high correlation between the features also made the classification more difficult.
On one hand, the Minimum Redundancy Maximum Relevance (MRMR) feature selection algorithm helped us \roy{get} to a set of measurable multimodal features sufficient to detect intention of interaction towards a robot based on human selected features set.
On the other hand, trying to deals with all 99 available features to elicit the more relevant ones fails.
New experiment\roy{s} need to be conducted with more variability in order to \roy{improve the scores of the intention detector using the reduced feature space.}

\doms{
Current work about high-level features fusion and analysis is ongoing.
}
We want to remove some artifacts that penalize feature selection and classification algorithms.
For instance, we are combining face detection, skeleton tracking and depth data to improve feature association.
These features now belong to the same user, i.e. when we compute Schegloff and face features they refer to the tracked pedestrian.
As far as we can say from our preliminary experimentation, doing so, results are improved but not significantly for the multi-user scenario.

\subsection{Impact of this research}

With this research we provide deeper knowledge about meaningful features that can facilitate robot's social abilities.
We computed new features inspired from the literature in social sciences, notably the Schegloff's features.
A ranked list of 32 most relevant features for our starting engagement can be found in \ref{sec:OrderedFeatures}.
This presented work gives design guidelines and praise multi-modal sensor embedding on robot to facilitate human-robot interaction.

The feature selection process depicted in this article was inspired from research on genome \cite{mundra2010svm}.
Principal Component Analysis (PCA) and the Linear Discriminant Analysis (LDA) did not provide significant classification improvement.
The MRMR algorithm however showed that selecting features can be more relevant than combining them.
The direct ranking provides information about meaningful features for
a classification task, for example.
The feature selection method can be applied in many contexts where space reduction is of interest.

This work is one more step into the use of multimodality for social signal processing applied to human-robot interaction.
Multimodality can be \roy{very useful in decoding and recognising} affect signals and hence \roy{in improving} the human-robot relationship.
With more and more powerful embedded system\roy{s deployed} on robots, we can expect such multimodal detection to be generalized in real-time and to allow robot\roy{s} to predict intentions of the users.
The prediction of the engagement is a first step toward\roy{s} a smoother and socially acceptable human-robot interaction.

\section{Acknowledgments}
The author would like to thank the Robosoft company for loaning us the \kompai{} robot, Inria and French Ministry of Education and Research for their support.

\section{References}
\bibliography{library}

\newpage
\appendix
\section*{APPENDIX}
\setcounter{section}{1}

\subsection{Samples of the corpus}
\label{sec:sample}
The table \ref{tab:kinectv} shows samples of the dataset that we propose for 4 of the events \cwinter{}, \cinter{}, \clinter{} and \csomeone{}. As one can see the event \clinter{} can be quite confusing with the \cinter{} state.

\begin{table}[H]
\centering
\begin{tabular}{c}
 \cwinter{} \\
 \includegraphics[width=0.85\linewidth]{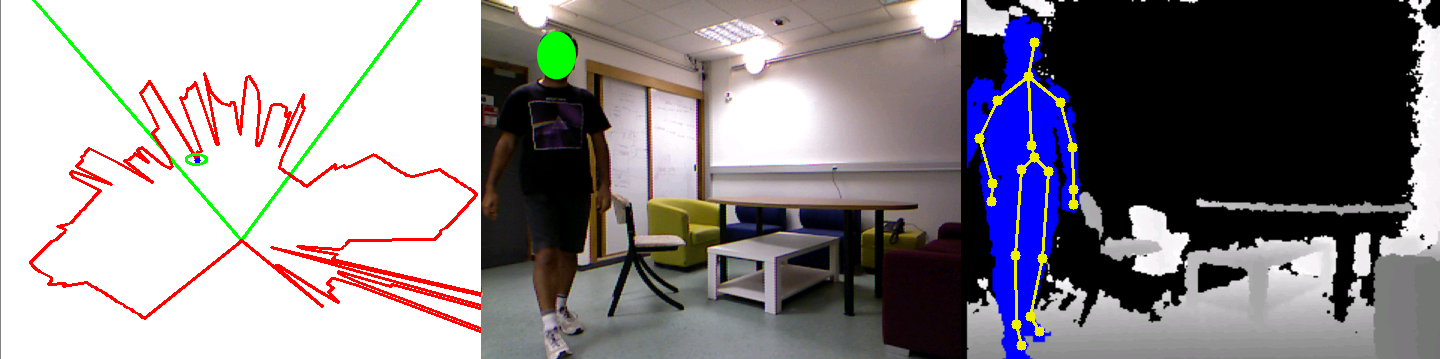} \\
 \hline \\
 \cinter{} \\ 
 \includegraphics[width=0.85\linewidth]{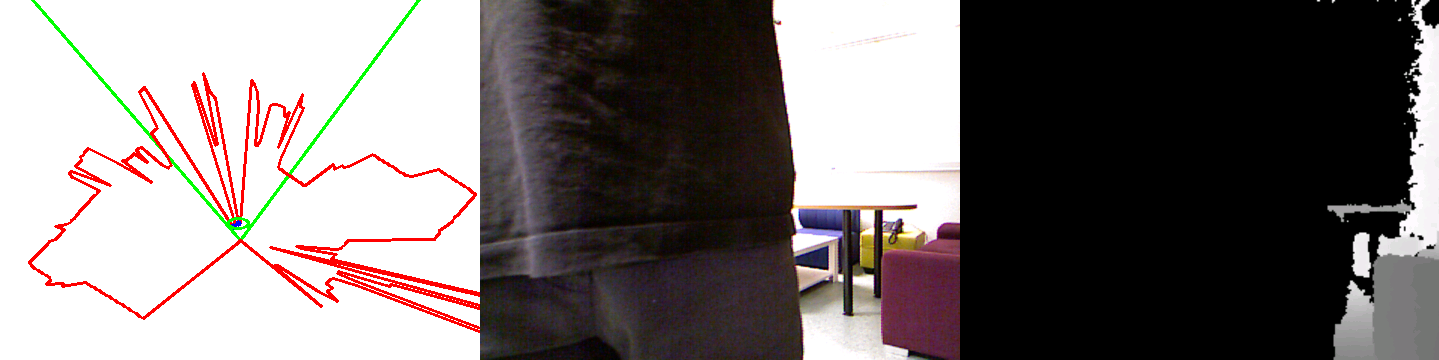} \\
 \hline \\
 \clinter{} \\
 \includegraphics[width=0.85\linewidth]{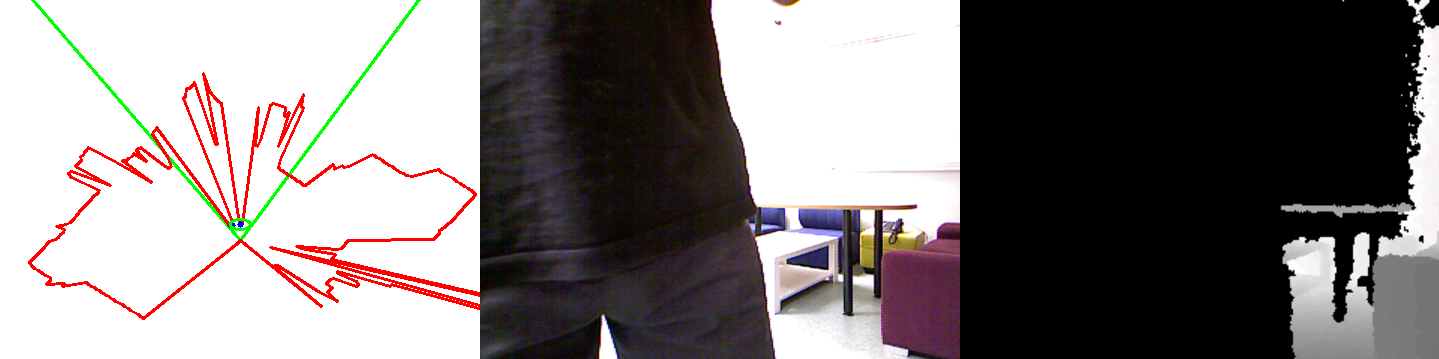} \\
 \hline \\
 \csomeone{} \\
 \includegraphics[width=0.85\linewidth]{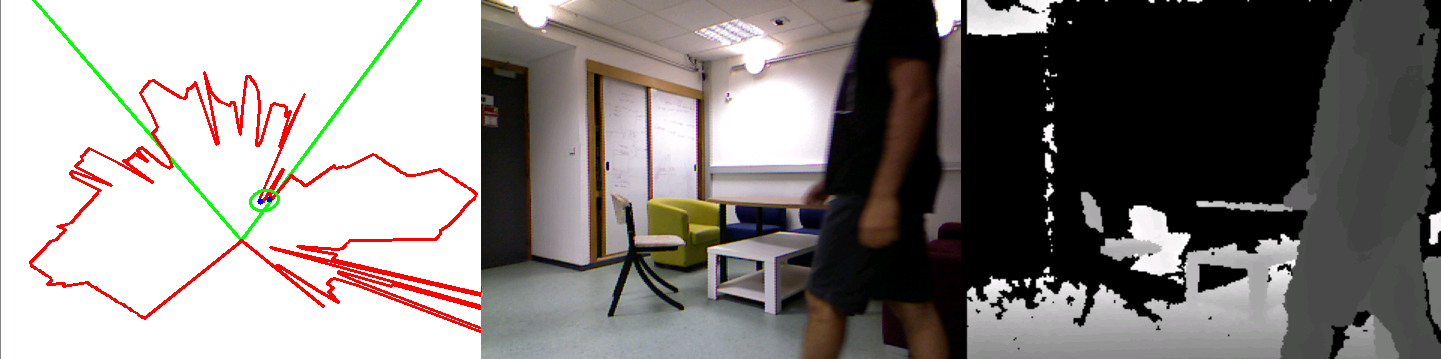} \\
\end{tabular}
\caption{Samples of data recorded with the \kompai{} equipped with a Kinect sensor and a laser telemeter.
For each view, one can find spatial information at left, rgb camera view with face detection in the middle and depth camera with people and skeleton detection.}
\label{tab:kinectv}
\end{table}

\subsection{Ordered list of 32 most relevant features}
\label{sec:OrderedFeatures}

Using the Minimum Redundancy Maximum Relevance (MRMR) algorithm, we ranked 32 features (see section \ref{sec:MRMR}).
In the following table, the reader must note that \textit{cible\_} prefix identifies pedestrian related features.
Other body features are related to Schegloff work (see section \ref{sec:bodypose}).
Features not listed in table \ref{tab:OrderedFeatures} are depicted in section \ref{chap:theoritical-approach}.

\begin{table}[htbp]
  \centering
  \caption{MRMR algorithm output on the 32 features set. }
    \begin{tabular}{ p{1cm} p{2.5cm} p{3cm} p{5cm} }
    \textbf{Order} & \textbf{Short name} & \textbf{Unit} & \textbf{Description}\\
\hline \\
    1     & shoulderPose\_rot & $radian$ & Rotation of the shoulder\\
    2     & cible\_dx & $meter.seconde^{-1}$ & Speed in x of pedestrian\\
    3     & cible\_y & $meter$ & position on Y axis of pedestrian \\
    4     & face\_size & $pixel$ & Size of face in the RGB frame \\
    5     & face\_x & $pixel$ & Lateral position of the face \\
    6     & beam  & $radian$ & Activated audio beam\\
    7     & angle & $radian$ & Audio localization (azimut) \\
    8     & hipPose\_x & $meter$ & Hip X attribute\\
    9     & hipPose\_y & $meter$ & Hip Y attribute \\
    10    & hipPose\_rot & $radian$ & Hip rotation angle \\
    11    & face\_y & $pixel$ & Height of the face \\
    12    & sad\_event & Speech/Not speech & Speech activity detection tags\\
    13    & stancePose\_rot & $radian$ & Stance rotation \\
    14    & torsoPose\_rot & $radian$ & Torso rotation \\
    15    & shoulderTorque & $radian$ & Shoulder torque \\
    16    & shoulderPose\_y & $meter$ & Shoulder Y attribute\\
    17    & source\_confidence & $[0;1]$ & Audio localization confidence\\
    18    & torsoTorque & $radian$ & Torso torque \\
    19    & stancePose\_z & $meter$ & Stance Z attribute \\
    20    & skl\_dist & $meter$ & Distance of the tracked skeleton \\
    21    & cible\_x & $meter$ & Position on X-axis of pedestrian \\
    22    & hipTorque & $radian$ & Hip torque \\
    23    & torsoPose\_y & $meter$ & Torso Y attribute\\
    24    & torsoPose\_x & $meter$ & Torso X attribute\\
    25    & shoulderPose\_x & $meter$ & Shoulder X attribute \\
    26    & stancePose\_x & $meter$ & Stance X attribute \\
    27    & cible\_dy & $meter.seconde^{-1}$ & Speed on Y-axis of pedestrian \\
    28    & cible\_dist & $meter$ & Distance of the pedestrian \\
    29    & torsoPose\_z & $meter$ & Torso Z attribute \\
    30    & stancePose\_y & $meter$ & Stance Y attribute \\
    31    & hipPose\_z & $meter$ & Hip Z attribute \\
    32    & shoulderPose\_z & $meter$ & Shoulder Z attribute \\
    \end{tabular}%
  \label{tab:OrderedFeatures}%
\end{table}%

\end{document}